# Social Media as an Instant Source of Feedback on Water Quality

Khubaib Ahmad[1], Muhammad Asif Ayub[1], Kashif Ahmad[2], Jebran Khan[4], Nasir Ahmad[1], Ala Al-Fuqaha[2]

1 Department of Computer Systems Engineering, University of Engineering and Technology, Peshawar, Pakistan.
2 Department of Computer Science, Munster Technological University Cork, Ireland.
3 Information and Computing Technology (ICT) Division, College of Science and Engineering (CSE), Hamad Bin Khalifa University, Doha, Qatar.
4 Department of Artificial Intelligence, Ajou University, Suwon 16499, Republic of Korea.

*Abstract*— This paper focuses on an important environmental challenge: water quality by analyzing the potential of social media as an immediate source of feedback. The main goal of the work is to automatically analyze and retrieve social media posts relevant to water quality with particular attention to posts describing different aspects of water quality, such as watercolor, smell, taste, and related illnesses. To this aim, we propose a novel framework incorporating different preprocessing, data augmentation, and classification techniques. In total, three different Neural Networks (NNs) architectures, namely (i) Bidirectional Encoder Representations from Transformers (BERT), (ii) Robustly Optimized BERT Pre-training Approach (XLM-RoBERTa), and (iii) custom Long short-term memory (LSTM) model, are employed in a merit-based fusion scheme. For merit-based weight assignment to the models, several optimization and search techniques are compared including a Particle Swarm Optimization (PSO), a Genetic Algorithm (GA), Brute Force (BF), Nelder-Mead, and Powell's optimization methods. We also provide an evaluation of the individual models where the highest F1-score of 0.81 is obtained with the BERT model. Overall, in merit-based fusion, better results are obtained with BF achieving an F1-score score of 0.852. We also provide a comparison against existing methods, where a significant improvement for our proposed solutions is obtained. We believe such a rigorous analysis of this relatively new topic will provide a baseline for future research.

## I. INTRODUCTION

OVER the last decade, social media outlets have been proven an effective source of communication and information dissemination. Their capabilities to engage large volumes of audience worldwide make them a preferred platform to discuss and convey concerns over different domestic and global challenges [1], [2]. The literature already reports their effectiveness in a diversified set of societal, environmental, and technological topics, such as food security [3], discrimination and racism [4], hate speech and crime [5], [6], public health [7], natural disasters [8], and technological conspiracies [9].

There have been also debates and discussions on air and water quality in social media outlets. In such discussions, generally, different topics, such as bad taste, color, and smell of drinking water, potential causes of the pollution, its impact on public health, and associated diseases, are discussed. It is noticed that, in their posts, social media users generally explicitly identify the regions having water-related issues along with the relevant information. For instance, public authorities can use this information as valuable feedback authorities on water distribution networks. However, generally, several challenges are associated with the extraction of relevant information from such informal sources. For instance, it is possible that the posts containing relevant or similar keywords do not represent the actual debates on water quality. Manually filtering and analyzing large collections of social media posts is a tedious and time-consuming process.

Recently, Machine Learning (ML) and Natural Language Processing (NLP) techniques have shown outstanding capabilities in similar applications. We believe ML and NLP techniques could also be employed in this interesting important application by automatically analyzing and filtering water-related social media posts. The automatic acquisition/retrieval and analysis of the water quality analysis techniques will result in a significant reduction in time and efforts spent in the process. To explore the potential of ML and NLP techniques in the domain, in this work, we propose a classification framework incorporating several pre-processing, data augmentation, classification, and fusion techniques. The pre-processing techniques allow cleaning the data by removing URLs and punctuation etc. The data augmentation, where we used a back translation scheme, serves two purposes. It not only increases our training set but also helped to balance the dataset by increasing the number of training samples in minority classes. For fusion, we employed both the naive fusion method by treating all the models equally as well as merit-based fusion schemes. In the merit-based fusion, five different weights selection/optimization techniques are used to assign weights to three different state-of-the-art architectures; namely, BERT, XML-RoBERTa, and LSTM, based on their performances. We also provide an evaluation of the individual models. We believe such rigorous analysis of the relatively new application will provide a baseline for future research.

The main contributions of the work can be summarized as follows.

- We explore a relatively new application; namely, water quality analysis in social media posts by proposing a

- complete framework starting from pre-processing, data augmentation, classification, and fusion techniques.
- We evaluate multiple state-of-the-art models both individually and jointly using several late fusion techniques.
- We also demonstrate how the performance of the classification framework improves by considering the performance of individual models in a late fusion scheme using five different weight selection and optimization techniques.

The rest of the paper is organized as follows: Section II provides an overview of the related work. Section III provides a detailed description of the proposed methodology and fusion techniques employed in this work. Section IV provides the description of the dataset, evaluation metrics, and experimental results. Finally, Section V concludes the work.

## II. Related Work

Being an important factor in a healthier life, water quality monitoring and analysis have always been a concern for public authorities. To this aim different strategies and sources of information, such as satellite imagery and crowd-sourcing platforms [10], are utilized. For instance, Galvin et al. [11] proposed a mobile application; namely, Cyanobacteria Assessment Network (CyAN) to monitor, detect, and disseminate information about water quality in lakes using remotely sensedimages. Similarly, Mohsen et al. [12] employed remote sensingtechniques to monitor and analyze the quality of the water of *Lake Burullus* in Egypt in satellite imagery. Satellite imagery has also been employed in several other interesting works for water quality analysis in lakes in different parts of the world [13], [14]. However, satellite imagery provides a bird's eye view and is generally used for large water reservoirs, such as lakes and dams.

On the other hand, crowd-sourcing techniques help in obtaining more detailed, contextual, and localized information. The literature already reports several interesting crowdsourcing solutions for monitoring water quality. For instance, Rapousis et al. [10] proposed QoWater, a client-to-server architecture-based mobile application allowing mobile users to give feedback on water quality. Similarly, Jakositz et al. [15] proposed and conducted a competition-based crowdsourcing study for tap water quality monitoring. One of the main drawbacks of such platforms for crowdsourcing is the limited number of users.

This limitation could be addressed with social media. Social media outlets, such as Facebook, Twitter, and Instagram, provide access to a large number of users. Such platforms could be utilized to provide instant feedback on water quality [16], [17]. However, extracting meaningful information from such informal sources is very challenging.

Thanks to the recent advancement in ML and NLP techniques, social media information could be automatically analyzed and filtered to extract relevant information. There are already some efforts in this direction. For instance, Lambert [18] performed sentiment analysis on users' feedback in social media posts to obtain their opinion and perception of tap water quality. Li et al. [19], on the other hand, provide a sentiment analysis of public opinions on social media on recycled water in China. Similarly, sentiment analysis of social media posts is also carried out by Do [20] where some basic NLP techniques, such as Bag of Words (BoW) and Naive Bayes, Bernoulli Naive Bayes, and Logistical Regression classifiers are used.

More recently, water quality analysis in social media posts has also been introduced in a benchmark competition; namely, MediaEval [21]. In the task, participants were asked to develop automatic tools that differentiate between relevant (i.e., water quality discussions) and irrelevant Twitter posts. The task mainly focuses on tweet text, however, additional information in the form of images associated with the tweets and metadata. In total, two teams, including our team, completed the tasks by introducing several interesting solutions. For instance, Hanif et al. [22] proposed a multimodal solution incorporating both images and text. For visual content, a pre-trained model namely VGGNet is fine-tuned while for textual features aBERT model is fine-tuned on the dataset. The authors also submitted the results of the visual information-based solution only. However, the results indicate lower performances for both solutions. On the other hand, considering the quality of the textual and visual content, our team [23] decided to focus on textual information only. To this aim, three different NNs models; namely, BERT, RoBERTa, and a custom LSTM, are employed both individually and jointly in a naive fusion scheme by treating all the models equally.

However, we believe merit-based fusion schemes could better exploit the potential of the models by assigning weights to the models based on their performance.

## III. Proposed Methodology

Figure 1 provides the block diagram of the proposed methodology. The proposed method can be roughly divided into three phases starting with a pre-processing and data augmentation phase where several strategies are used to clean and augment the data. After pre-processing, multiple NNs models are trained on the data. Finally, the classification scores obtained with the individual models are combined using several merit-based late-fusion schemes. In the next subsections, we provide a detailed description of each phase.

### A. Pre-processing and Data Augmentation

In the pre-processing and data augmentation phase, we employed different strategies to clean and increase the number of training samples. As a first step, we cleaned the text by removing URLs, account handles, emojis, and unnecessary punctuation. After cleaning the data, we performed data augmentation using text translation technique; namely, back-translation. The text translation approach naturally suits our application as the dataset is composed of both Italian and English tweets where Italian tweets are translated into English and added to the original training set. This technique helps in generating more training samples without disturbing the context of the data.

Data augmentation serves two purposes. Firstly, it increased our training set. Secondly, it helped in balancing the training set by increasing the minority class. Moreover, along with data augmentation, we also used an up-sampling technique to balance the dataset.

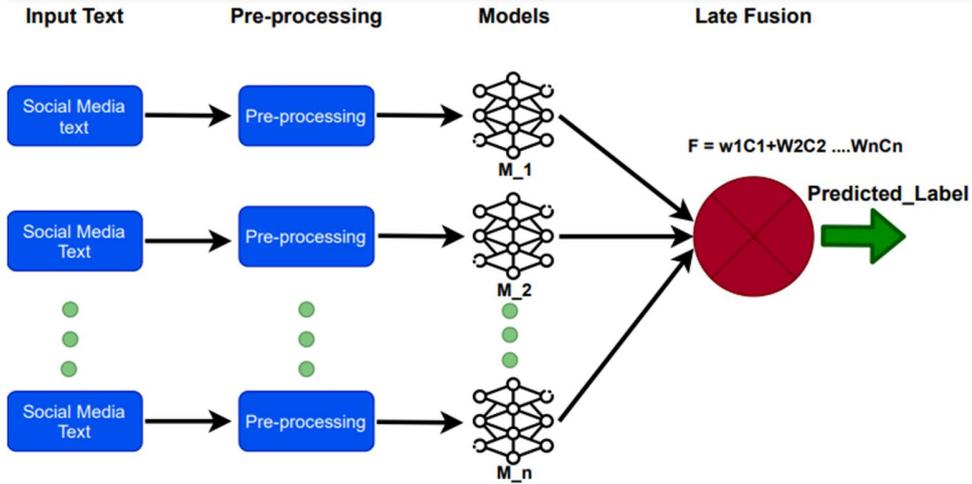

Fig. 1: Block diagram of the proposed methodology.

### B. Model Training and Classification

In this work, we used three different state-of-the-art Neural Networks (NNs) architectures; namely, BERT [24], XLM-RoBERTa [25], and LSTM. In the case of BERT and XLM-RoBERTa, we fine-tuned existing pre-trained models while a customized model is trained for LSTM. The details of the models are provided below.

- **BERT-based Solution**: BERT is a multi-layer encoder. In contrast of conventional NLP models, BERT relies on bi-directional training mechanism by taking into account both the previous and next tokens. Such training capabilities allow it to better extract contextual information. In this work, we rely on a pre-trained BERT model, which is fine-tuned on the water quality analysis dataset. The model is composed of 12 layers, 12 attention heads, and 110 million parameters. We note that necessary pre-processing of the cleaned training data is carried out, using *Tensorflow* libraries, to bring the data in the required form to be fed into the model. Since our dataset is composed of two classes only (i.e., binary classification task), we used the *Binary Cross entropy* loss function along with the *Adaptive Moments (Adam)* optimizer. Table I summarizes the parameters setting of the BERT model used in this work.

TABLE I: Parameters setting of the BERT model.

| Attribute | Value |
|---|---|
| Features | 768 |
| Number of Layers | 12 |
| Fully Connected | 2 (fully connected feed-forward layers) |
| Cost Function | Binary crossentropy |
| Number of Classes | 2 |
| Training Solver | Adam |
| Mini-Batch Size | 32 |
| Dropout | 0.1 |
| Validation data | yes |

- **XLM-RoBERTa-based Solution**: XLM-RoBERTa is a multilingual version of RoBERTa. RoBERTa itself is a modified version of BERT. The model is trained on a large-scale dataset covering text from 100 different languages using Mask Language Modeling (MLM) objective. The multi-lingual nature of the model makes it a better choice for our application.
  As per the requirements of the model, the input text is tokenized before feeding into the model. The model is then fine-tuned on the water quality analysis dataset using an Adam optimizer with a binary cross-entropy loss function. A summary of the parameter settings of the model is provided in Table II.

TABLE II: Parameters setting of the RoBERTa model.

| Attribute | Value |
|---|---|
| Features | 1024 |
| Number of Layers | 36 |
| Cost Function | binary cross-entropy |
| Number of Classes | 2 |
| Training Solver | Adam |
| mini-batch size | 32 |
| Dropout | 0.1 |
| Epochs | 20 |
| Token Max Length | 512 |
| Validation data | yes |

- **LSTM-based Solution**: Our third model is based on LSTM architecture. LSTM is a Recurrent Neural Network (RNN) with better memorizing pattern capabilities, which makes it a better choice for text classification compared to classical ML algorithms, such as decision trees, Random Forests (RF), and Support Vector Machines (SVMs). In this work, we used a customized model composed of three layers including an input, LSTM, and output layer. Our model is composed of 491,713 trainable parameters. A summary of the parameter settings of the model is provided in Table III.

TABLE III: Parameters setting of the LSTM model.

| Attribute | Value |
|---|---|
| Total Layers | 4 |
| Embedding Vector Space | 32 |
| LSTM Vector space | 64 |
| Cost Function | crossentropy |
| Number of Classes | 2 |
| Training Solver | Adam |
| Mini-Batch Size | 64 |
| Activation | Sigmoid |
| Drop out | 0.1 |
| Learning Rate | 0.001 |
| Epochs | 20 |
| Validation data | yes |

### C. Fusion

For the fusion, we mainly rely on late fusion schemes where both naive and merit-based fusion methods are employed. Our baseline method is based on a simple aggregation of the classification scores obtained with all the models. In merit-based fusion, we deploy different optimization and search techniques to optimize the weights assigned to the models in fusion. We note that in the current implementation, we use a linear combination of the models in the late fusion using Equ. 1.

$$F_c = W_1 C_1 + W_2 C_2 + ... + W_n C_n \quad (1)$$

Here $F_c$ represents the combined classification score obtained as an outcome of fusion while $w_n$ denotes the weight assigned to the nth model whose score is represented by $C_n$. In our case, $n = 3$.

In the case of naive baseline fusion, all the models are assigned equal weights (i.e., $W_1 = W_2 = W_n = 1 = 1/N$). In the merit-based fusion, on the other hand, values of the weights are selected based on the optimization/search methods used in this work. The details of the fusion methods used in this work are provided below.

*1) Particle Swarm Optimization (PSO) Based Fusion:* The use of PSO based method is motivated by its promising performance in similar tasks [26]–[28]. The key concept of PSO is inspired by the social behavior of birds flocking and fish schooling, where the idea is to get benefit from the experience of each other in finding the best solution. To this aim, it starts with an arbitrary population of the possible solutions, which are termed as particles, and tries to iteratively optimize the potential solutions to satisfy a given constraint provided in the objective function. To find the best global minimum, the algorithm keeps track of the current and best position and velocity of each particle at each iteration, which are then updated in successive iterations.

In this work, each combination of weights is considered a potential solution whereas our objective function is based on the accumulative error ($E_{acc}$), computed by Equ. 2.

$$E_{acc} = 1 - A_{acc} \quad (2)$$

Here $A_{acc}$ represents the cumulative accuracy computed on the validation set using Equ. 3. In the equation, $p_n$ represents the probability/score obtained with the $n^{th}$ model while $x(n)$ is the weight to be assigned to the $n^{th}$ model.

$$A_{acc} = x(1) * p_1 + x(2) * p_2 + ... + x(n) * p_n \quad (3)$$

It is important to mention that PSO is a heuristic solution, and the solution is not guaranteed to be optimal. However, the literature indicates that the solutions found by PSO are generally close to the optimal one. Table IV provides a summary of the pros and cons of PSO.

*2) Genetic Algorithm Based Fusion:* Genetic Algorithm, which is inspired by Charles Darwin's theory of natural evolution, has also been widely explored in the literature for similar tasks involving weight selection and optimization [26]. The basic idea behind GA is to incorporate the natural evaluation phenomena in search/selection problems by selecting the best one among the potential solutions at each iteration.

The GA-based search/selection process is composed of several phases. The process starts with a random population of individuals/potential solutions (i.e., randomly selected weight combinations). The algorithm then searches for the fittest individuals (i.e., weight combination) by evaluating the individuals/potential solutions against fitness criteria provided in the fitness function, iteratively. The process continues by employing crossover and mutation operations until the population convergences (i.e., no further improvement is possible). Crossover and mutation are key operations directly contributing to the performance of the algorithm. The former aims to push the population towards local minimum/maximum and the latter aims to explore the best among the candidate local minimum/maximum solutions.

In this work, similar to PSO-based fusion, our fitness criterion is based on the accumulated error ($E_{acc}$), which is computed on a separate validation set using Equ. 2.

*3) Brute Force Based Fusion:* The BF search, which is also known as an exhaustive search, tries all the possible solutions to find a satisfactory one for an underlying problem. The BF method brings several advantages. For instance, it guarantees the best solution by considering all the possible solutions before choosing the best one. Moreover, a simple working mechanism makes it a preferred solution for small problems in a diversified list of applications. However, on the other hand, there are several limitations of the method. For instance, the complexity of the method is very high in high-dimensional applications. Moreover, it takes a lot of time to find all the possible solutions for a high-dimensional application.

In this work, we used the method to find the best combination of the weights assigned to the models that minimize the classification error. This method suits our task as we have only three models and the method needs to find very few combinations for the selection of the best combination. In this work, we used an open-source Python library, namely, SciPy for the implementation of the algorithm, which aims to find the grid point having the lowest value of the objective function (i.e., cumulative error defined by equation 2).

*4) Powell's Method Based Fusion:* In this method, we rely on the evolutionary Powell's method for the optimization of the weights to be assigned to the individual models during fusion. The method is inspired by the original method [29], however, a stochastic element is introduced. The method aims at the global minima of the objective function, which is based on a cumulative error in our case.

The algorithm works in several steps. The process starts by randomly selecting and evaluating a few points/solutions. In the second step, a list of parameters is selected in random order. In the third step, a portion of the previously evaluated points/solutions is used as parents by ensuring the selection of points with the lowest error. The algorithm then looks for the children in the next generation. Finally, all the children are evaluated in the fifth step and the process is repeated from the third step. It is important to mention that, while searching for the minimum, the algorithm moves in one direction only until it finds the local minima. The algorithm moves in the other direction once the minimum is found in the current direction.

Similar to the other methods used in this work, our objective/cost function is based on the cumulative error defined by Equ. 2. For the implementation of this method, we used a Python open-source library, namely, SciPy.

*5) Nelder–Mead Based Fusion:* Nelder-Mead algorithm, also known as a pattern search, is considered one of the suitable algorithms for both one-dimensional and multidimensional optimization problems [30]. The algorithm is based on the concept of a simplex (i.e., a special polytope of $n+1$ vertices in $n$ dimensions). This implies that the algorithm produces and keeps a set of $n+1$ dimensions for an $n$-dimensional task. The algorithm then computes the values of the objective function for each point to find and replace one of the oldest points with a new one, iteratively. In our case, $n = 3$, and the objective function is based on the cumulative error defined by Equ. 2. For the implementation of the method, we used a Python open-source library, namely, SciPy.

## IV. RESULTS AND ANALYSIS

### A. Dataset

For the evaluation of proposed solutions, we used a large-scale dataset introduced in a benchmark competition task namely "WaterMM: Water Quality in Social Multimedia" MediaEval 2021 [21]. The dataset is composed of a large collection of Twitter tweets tweeted in English and Italian from May 2020 to April 2021. The data is collected using English and Italian keywords related to water quality, color, pollution, and water-related illnesses. The main challenge lies in differentiating between water quality-related tweets and irrelevant tweets containing terms, such as water, floods, etc. Table ?? provides some sample tweets from both classes.

The dataset is manually annotated by analyzing the tweets under the guidelines provided by the authorities of the Eastern Alps River Basin District, who are responsible for hydrogeological defense in North-East Italy. Each tweet is annotated as either relevant or irrelevant to water quality. In total, the dataset is comprised of 10,000 tweets. The training and test sets are already separated by the task organizers. The training set is comprised of 8,000 tweets and the test set covers 2,000 tweets. The training set is imbalanced containing only around 17.18% of relevant tweets, which poses challenges in training AI models.

It is important to mention that the dataset also provides some images associated with tweets. However, images are associated with very few tweets. A total of 960 images are covered in the dataset, where the majority of the images are irrelevant to the task. Figure 2 provides some sample images from the dataset. Considering the quantity and quality of the images, we did not consider visual contents in our experiments.

### B. Evaluation Metrics

The evaluation of the proposed solutions is carried out in terms of three different metrics: namely, precision, recall, and F1-score. These metrics are also used in the benchmark competition where the dataset was introduced.

### C. Experimental Results

In this section, we provide a detailed description and analysis of the experimental results. We also provide comparisons against the methods proposed in the benchmark competition.

*1) Evaluation of the Individual Models:* Table VI provides the experimental results of the individual models in terms of precision, recall, and F1 score. As can be seen, overall better results are obtained with the BERT model while surprisingly the least score is obtained with RoBERTa. However, there is no single winner in terms of all of the three metrics. For instance, the highest F1 score is obtained with the BERT model while LSTM achieved the highest score in terms of recall. The lower F1-Score for LSTM compared to BERT indicates a higher number of false-positive samples. This variation in these metrics is an indication of the variations in the performances of these models in both classes, which provides the basis for our fusion-based experiments.

*2) Evaluation of the Fusion Methods:* Table VII provides the experimental results of the fusion experiment. In this experiment, we evaluated several fusion methods including a naive method by treating all the models equally and a merit-based fusion strategy where five different weight selection/optimization methods are employed to assign weights to the models.

As can be seen, interestingly, no significant improvement in the performance of the naive baseline fusion scheme over the best-performing individual model has been observed. One of the possible reasons could be the low-performing model (i.e., XML-RoBERTa) as it simply aggregates the classification scores of all the models. These results provide the basis for our merit-based fusion where weights are assigned to the models based on their performance in the first experiment.

In merit-based fusion, significant improvement in terms of F1-score is obtained with all the methods over the best performing individual model and the naive baseline fusion method. This emphasizes the fact that merit-based weights should be assigned to contributing models in the fusion.

As far as the performance of the individual weights selection strategies for the merit-based fusion is concerned, better results are obtained with BF based method having an improvement of .001 and .009 over the Powell and PSO-based fusion methods, respectively. One of the potential causes of better results of BF is its ability to provide a guaranteed best solution, where it first searches for all possible combinations and pick

TABLE IV: A summary of the pros and cons of the weights selection/optimization methods used in this work.

| Method | Pros | Cons |
|---|---|---|
| PSO | - Easy to understand and implement<br>- Efficient and insensitive to scaling<br>- Fewer parameters<br>- Suitable for concurrent parameters | - May easily fall and stuck into local minima<br>- Low convergence rate<br>- It is a heuristic solution and does not guarantee a globally optimalsolution |
| GA | - Robust to local minima<br>- Can be easily parallelized for concurrent processes | - Computationally expensive |
| BF | - Provides a guaranteed best solution<br>- Applicable to several problems from different domains<br>- Simple to understand and implement<br>- Better suited for small problems | - Its complexity increases with an increase in the dimensionality of the problem<br>- Very slow |
| Powell's Method | - The objective function does not need differentiable | - May not find local minima in many iterations |
| Nelder–Mead | - Works with function evaluations only | - Not efficient<br>- May take a large number of iterations without many changes in function value |

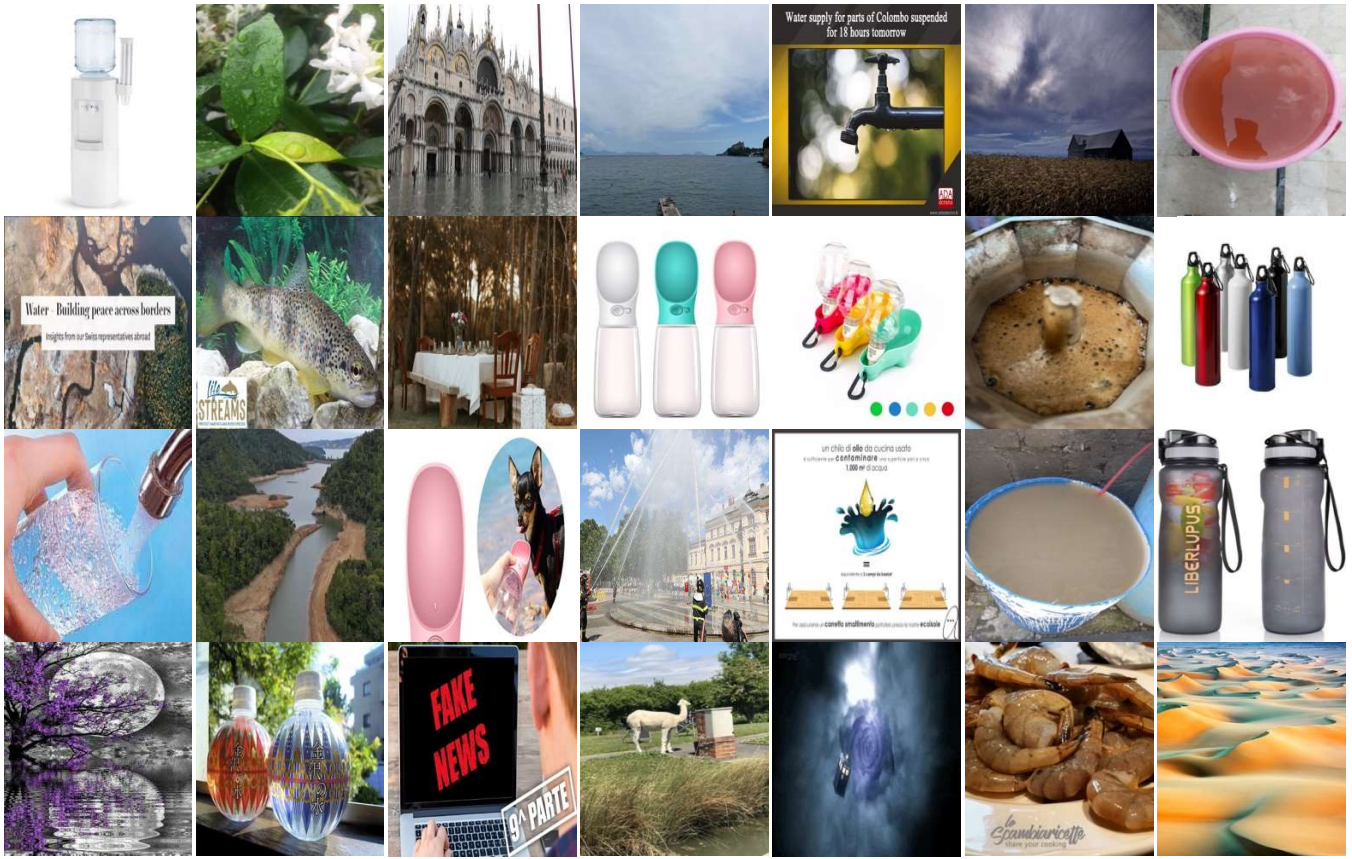

Fig. 2: Sample images from the dataset.

TABLE V: Some sample tweets from the dataset

| Relevant Samples | Irrelevant Samples |
|---|---|
| "Alta concentrazione di cloro, acqua non potabile nella zona di San Pietro Lametino" (**Translation:** High concentration of chlorine, non-drinking water in the San Pietro Lametino area) | "Neanche le 7 ed ho giÀ. cambiato una gomma ad un collega sul raccordo sotto l'acqua". (**Translation:** Not even 7 and I have already changed a tire to a colleague on the fitting under the water.) |
| "PPPP Shame please supply water #TharNeedsWaterCanal" | "Someone just threw a water bottle towards PPD". |
| "1 out of every 3 people on our planet do not have access to clean water." | "SotDPodcast Need one for my water bottle!. Bottle your tears for me to water my plants latest by 9AM Monday." |
| "Se l'occidente avesse problemi di acqua potabile, non ci metteremmo sue secondi a depurare l' acqua del mare. lo stesso bisognerebbe fare con i paesi del sud del mondo." (**Translation:** If the West had problems with drinkingwater, it would not take us seconds to purify the sea water . the same shouldbe done with the countries of the southern hemisphere.) | "ho rovesciato una bottiglia d'acqua, mi sono fatta il bagno e allagato la cucina, tt bn." (**Translation:** I spilled a bottle of water, took a bath and flooded the kitchen, tt bn.) |
| "Almeno il 15% delle terre coltivate del Pianeta subisce una carenza idrica non dovuta a vincoli idrologici ma causata da unaâ C¨ ̂(**Translation:** At least 15% of the planet's cultivated land suffers from a water shortage not due to hydrological constraints but caused by a ...) | "a cena stasera ho un vegetariano e una che Ä¨ a dieta e non mangia carboidrati e legumi QUINDI mangeremo una bottiglia di acqua naturale." (**Translation:** for dinner tonight I have a vegetarian and one who is on a diet and does not eat carbohydrates and legumes SO we will eat a bottle of still water.) |
| "Neyyar water supply project faces more delays." | "frosty water bottle." |
| "Grazie al monitoraggio della gravitÀ. sulla Terra Ä¨ possibile stimare la situazione dell'acqua dolce a livello planetario e praˆ." (**Translation:** Thanks to the monitoring of gravity on Earth, it is possible to estimate the situation of fresh water at the planetary and praˆ C level.) | "Pessima non l'idea in se per noi ma la gestione, comunicazione e condivisione di una tale misura! Una tempesta in un bicchiere d'acqua a conferma di gravi problemi di comunicazione del nostro Governo." (**Translation**: Bad not theidea itself for us but the management, communication and sharing of such a measure! A storm in a glass of water confirming our government's serious communication problems.) |
| "RT @ContrattiFiume: Che cosa finisce nelle nostre #acque? E con quali impatti su #salute e #ambiente? Utilizzati come discariche dove smaltà." (**Translation**: RT ContrattiFiume: What ends up in our #waters? And with what impacts on #health and #environment? Used as landfills where glazedâ.) | "@ApolloVentuno Adesso? che meraviglia Qui siamo ancora sotto acqua a catinelle." (**Translation**: ApolloVentuno Now? wonderful! Here we are still underwater!) |
| "Acqua di nuovo potabile a Cabbio e Muggio https://t.co/iiVXhdj6iH." (**Translation**: Drinking water again in Cabbio and Muggio https://t.co/iiVXhdj6iH.) | "Naughty Doctor Ma ve li ricordate i commenti dopo l'acqua alta del 12 novembre? Il tasso di ignoranza altissimo e non conosce differenze geografiche." (**Translation**: Naughty Doctor But do you remember the comments after the high water on November 12? The rate of ignorance is very high and knows no geographical differences.) |
| "Serra, ordinanza del commissario: non potabile la ̂acqua proveniente dai serbatoi Scorciatinaâ. e Timpone Tondoâ. #Calabria #calabrianotizie #acqua." (**Translation**: Serra, order of the commissioner: the water from the aˆ Scorciatinaaˆ and Timpone Tondoaˆ tanks is not drinkable #Calabria #calabrianotizie #water.) | "all ever drunk a water bottle too fast and almost drown yourself." |

TABLE VI: Evaluation of the individual models in terms of precision, recall, and f1- score.

| Methods | Precision | Recall | F1-Score |
|---|---|---|---|
| BERT | 0.833 | 0.790 | 0.811 |
| XLM-RoBERTa | 0.81 | 0.579 | 0.687 |
| LSTM | 0.886 | 0.640 | 0.743 |

TABLE VII: Evaluation of fusion schemes in terms of micro-precision, recall, and f1-score.

| Methods | Precision | Recall | F1-Score |
|---|---|---|---|
| Baseline (Equal Weights) | 0.873 | 0.760 | 0.813 |
| PSO | 0.781 | 0.916 | 0.843 |
| GA | 0.791 | 0.895 | 0.840 |
| Brut Force | 0.810 | 0.900 | 0.852 |
| Powell's Method | 0.810 | 0.897 | 0.851 |
| Nelder–Mead Method | 0.862 | 0.807 | 0.834 |

TABLE VIII: Comparisons against existing methods in terms of micro-precision, recall, and f1-score.

| Methods | F1-Score |
|---|---|
| Hanif et al. [22] | 0.371 |
| Asif et al. [23] | 0.794 |
| This Work (PSO) | 0.843 |
| This Work (GA) | 0.840 |
| This Work (BF) | 0.852 |
| This Work (Powell's Method) | 0.851 |
| This Work (Nelder–Mead Method) | 0.834 |

the best weight combination that maximizes the performance. The literature suggests that the complexity of BF increases with an increase in dimensionality and consumes more computation power. However, in this work, we are consideringjust three models (i.e., dimension = 3) and thus the methodis better suited for this application in the current implementation. It is important to note that the difference in the performances of BF and other competing methods, suchas PSO, GA, and Powell's method is negligible.

*3) Comparison against Existing Solutions:* We also provide comparisons against existing works proposed for the task in a benchmark competition; namely, MediaEval 2021 [21]. In total, two teams managed to complete the task. Our teamalso participated in the competition and obtained the highest scores. As can be seen in Table VIII, all the merit-based fusion techniques employed in this work obtained significant improvements over the existing solutions. Our best performing merit-based method; namely, BF-based fusion, obtained an improvement of 48.1% over the method proposed by Hanif et al. [22]. On the other hand, it obtained an improvement of 4.1% over the method proposed by our team [23] for the task in the competition, where we proposed a naive fusion method by treating all the models equally.

The significant improvement in the performance of the water quality analysis framework indicates the significance of merit-based fusion.

*D. Lessons Learned*

The lessons learned during this work can be summarized as follows.

- Recently, water quality analysis got the attention of the research community, and several interesting solutions incorporating different sources of information have been proposed.

- Crowdsourcing has been one of the potential solutions to obtain relevant feedback on water quality, however, it is a tedious and time-consuming process to obtain a sufficient number of participants. Social media outlets provide a better platform to involve a large number of volunteers in crowdsourcing for water quality analysis.
- As demonstrated in this work, ML and NLP techniques allow to automatically analyze and extract relevant information from large collections of social media posts.
- The classification results are significantly improved by jointly employing the state-of-the-art models. However, individual models' performances need to be considered in assigning weights to the models in the fusion.
- The BF, though a computation-intensive method, obtained better results by searching all possible combinations of weights and choosing the one with the best results. However, in applications with fewer models (as we have in our application) the computational complexity is negligible.
- The difference in the performances of BF and other competing methods, such as PSO, GA, and Powell's method is negligible.
- The use of NLP and ML techniques will allow us to automatically acquire/retrieve and analyse the relevant posts containing citizens' complaints about drinking water as an additional source of information. This automatic analysis will result in a significant reduction in the efforts and time spent on manual feedback through crowd-sourcing techniques.

## V. Conclusions and Future Work

In this paper, we proposed an ensemble framework for water quality analysis in social media posts. To this aim, different preprocessing, data augmentation, classification, and fusion strategies are analyzed and evaluated. Though the social media posts contain images, we focused on textual information only. Our choice of using textual information only is mainly motivated by the quality and quantity of the images associated with the posts. Overall, we used three state-of-the-art NNs models individually and jointly in both naive and merit-based fusion methods. During the experiments, we observed better performance for merit-based fusion schemes, where weights were assigned to models based on their performance. This emphasizes the assumption that individual performances of the models should be considered in fusion.

In the future, we aim to incorporate the additional information available in the form of images and meta-data to further enhance the performance of the framework. However, the use of the additional information especially the images associated with social media posts is subject to the availability and quality of the images. To explore this aspect of the problem, we also aim to collect a collection of social media posts containing images along with the text. We believe, the additional information in the form of videos and images, for example showing the bad color of water, will complement the textual information.

Acknowledgment

This publication was made possible by NPRP grant # [13S-0206-200273] from the Qatar National Research Fund (a member of Qatar Foundation). The statements made herein are solely the responsibility of the authors.